\documentclass[letterpaper, 10 pt, conference]{ieeeconf}  

\IEEEoverridecommandlockouts                              

\usepackage{cite}
  \usepackage[dvips]{graphicx}
  \graphicspath{{./eps/}}
  \DeclareGraphicsExtensions{.eps}
\usepackage[tight,footnotesize]{subfigure}
\usepackage{url}

\title{\LARGE \bf
Evaluation of Plane Detection with RANSAC According to Density of 3D Point Clouds
}

\author{Tomofumi Fujiwara$^{1}$, Tetsushi Kamegawa$^{2}$ and Akio Gofuku$^{3}$
\thanks{$^{1}$Tomofumi Fujiwara, $^{2}$Tetsushi Kamegawa and $^{3}$Akio Gofuku are with
Graduate School of Natural Science and Technology, Okayama University, 
3-1-1 Tsushimanaka, Kita-ku, Okayama-shi, Okayama 700-8530, Japan
        {\tt\small $^{1}$fujiwara.t@mif.sys.okayama-u.ac.jp, $^{2}$kamegawa@sys.okayama-u.ac.jp, $^{3}$fukuchan@sys.okayama-u.ac.jp}}%
}

\begin{document}

\maketitle
\thispagestyle{empty}
\pagestyle{empty}

\begin{abstract}

We have implemented a method that detects planar regions from 3D scan data using Random Sample Consensus (RANSAC) algorithm to address the issue of a trade-off between the scanning speed and the point density of 3D scanning.
However, the limitation of the implemented method has not been clear yet.
In this paper, we conducted an additional experiment to evaluate the implemented method by changing its parameter and environments in both high and low point density data.
As a result, the number of detected planes in high point density data was different from that in low point density data with the same parameter value.

\end{abstract}


\section{Introduction}
It is necessary to save victims immediately who were trapped in damaged buildings when disasters happened.
Therefore many studies on rescue robots have been conducted recently.
The general issue of indirect-vision driving of a rescue robot arises due to the scenario that an operator can not approach to damaged buildings. 
In those cases, information gathered by a mobile robot should be displayed to an operator appropriately.

Some approaches to solve the problem are found in literatures.
Some of them use a Laser Range Finder (LRF) installed on a rotating stage to measure surroundings and display a 3D point cloud~\cite{surmann2003autonomous}. 
On the other hand, 3D scanning by using a LRF has a problem with the scanning speed and the point density due to the resolution of radial scan lines.
There is a trade-off between the scanning speed and the point density, which affects the operator's ability to recognize the environment.
We have addressed the issue of the trade-off by reducing the point density appropriately~\cite{fujiwara2013plane}.
We have already implemented a method that detects planar regions from 3D scan data using Random Sample Consensus (RANSAC) algorithm~\cite{fischler1981random}\cite{ransac_web} and integrates the overlapped planes and draws them as convex hulls using Graham's scan algorithm~\cite{graham1972efficient}.
We found that the same planes could be detected in both high and low point density data.

However, the characteristic and the limitation of the implemented method have not been clear yet.
In this paper, we conducted an additional experiment to evaluate the implemented method by changing its parameter and environments in both high and low point density data.
The parameter is a threshold which determines the number of points in a 3D point cloud regarded as a plane.
As a result, the number of detected planes in high point density data was different from that in low point density data with the same threshold.
Therefore, it is necessary to determine the threshold according to the point density.

\section{Hardware}
\begin{figure}[tbp]    
\centering
\includegraphics[width=3.5cm]{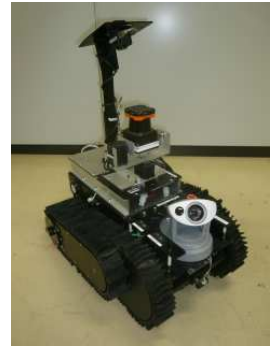}
\caption{The surveyor type robot which has a 3D laser scanner.}
\label{surveyor}
\end{figure}

A mobile robot shown in Fig. \ref{surveyor} is used as a component of our system. The robot is called as a {\it surveyor type robot}, which was developed by an R\&D project of the New Energy and Industrial Technology Development Organization (NEDO), Japan~\cite{journals/jfr/KamegawaSHUM11}. The size of this robot is 490(max:705)$\times$590$\times$400 cubic millimeters and its weight is 24 kilograms. It has a LRF mounted on a rotating stage and can perform 3D scan of the surroundings. 

\begin{figure}[tbp]
\begin{tabular}{ccc}
\begin{minipage}{0.45\columnwidth}
\centering
\includegraphics[width=1\columnwidth]{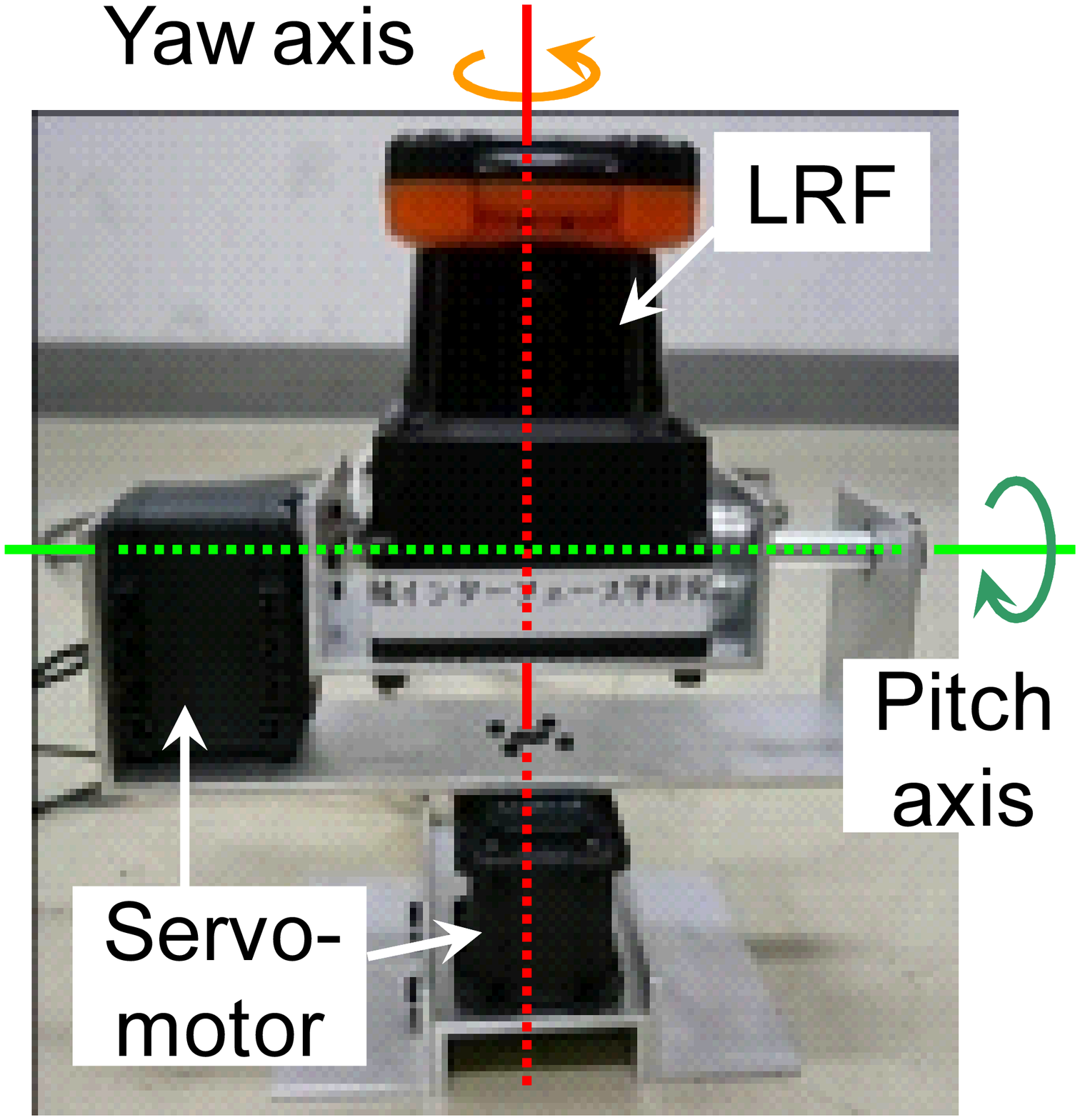}
\caption{2-DOF stage for 3D scan.}
\label{2dof_platform}
\end{minipage}

\begin{minipage}{0.05\columnwidth}
\end{minipage}

\begin{minipage}{0.5\columnwidth} 
\centering
\includegraphics[width=1\columnwidth]{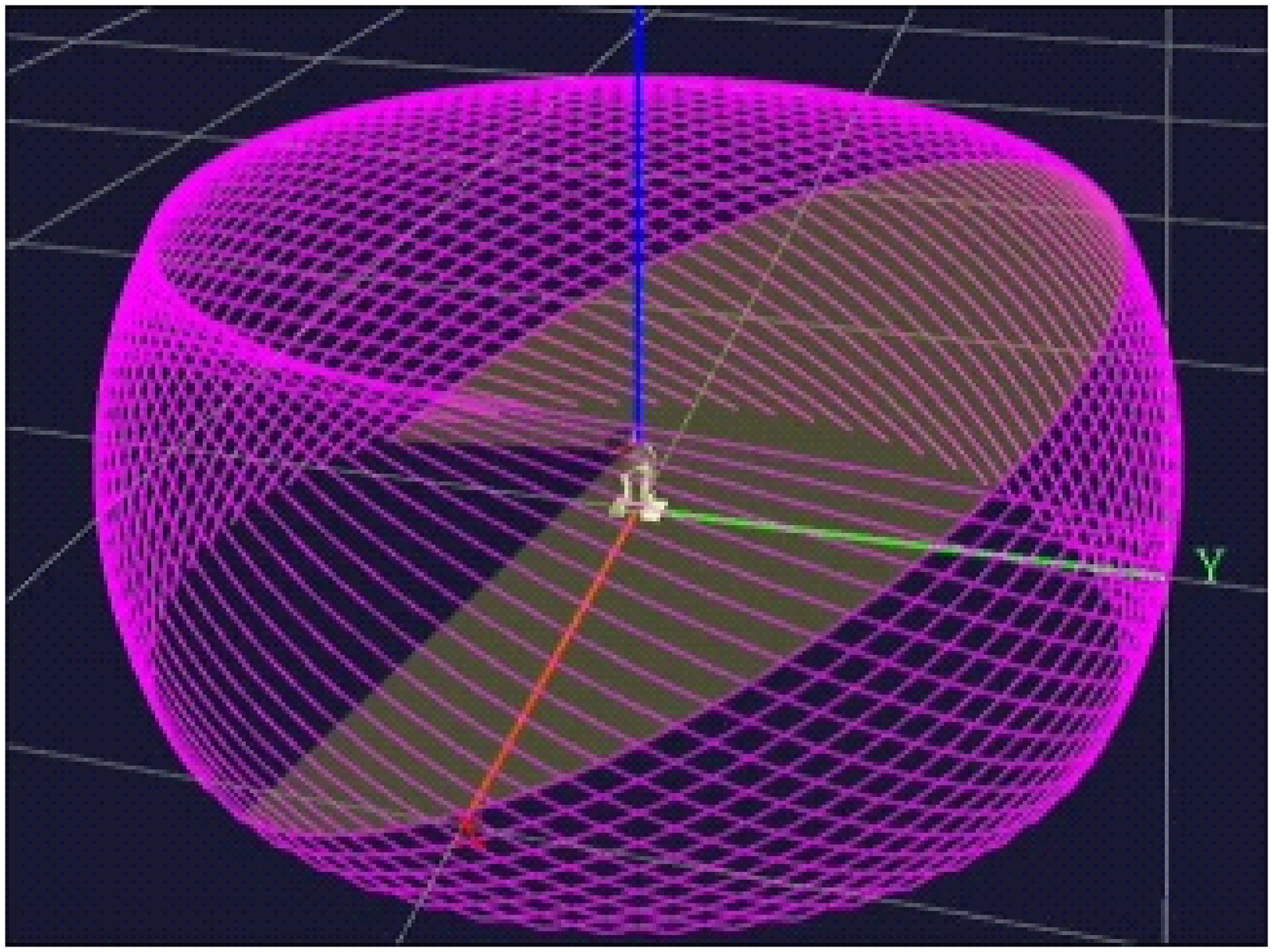}
\caption{Illustration of the range of 3D scan.}
\label{3d_scan}
\end{minipage}
\end{tabular}
\end{figure}

3D scan can be achieved by a rotating stage with two degrees of freedom (2-DOF stage) shown in Fig. \ref{2dof_platform}. HOKUYO UTM-30LX (Top-URG) is installed on it. This LRF is able to measure 2D distance data with the range up to 30 meters and 270 degrees' angle.
The minimum step angle of scanning is 0.25 degrees, thus 1,081 points are measured at one scan.
The stage has two servomotors (ROBOTIS Dynamixel AX-12) and performs 3D scan by tilting the scanning plane and then panning the LRF~\cite{ohno2009development}\cite{nagatani2008continuous}.
An illustration of a 3D scan range by using our system is shown in Fig. \ref{3d_scan}. 
In this paper, the pitch angle is fixed at 30 degrees and the yaw angle varies 0 to 300 degrees.
The point density of the data depends on a rotational step of the yaw angle, where the high point density data is obtained as 324,300 ($=300 \times 1,081$) points and the low point density data is obtained as 54,050 ($=50 \times 1,081$) points by six degrees' step to the yaw axis.

\section{Experiment and Results}
The plane detection method using RANSAC algorithm was conducted for 3D scan data with high and low point density.
Components to be estimated in RANSAC algorithm are normal vectors $\vec n$ that compose planes.
Parameters that can be configured in RANSAC algorithm are:
\begin{itemize}
\item $p_g$: The probability that a randomly selected point is part of a plane expressed by the estimated parameter $\vec n$. 
\item $p_{fail}$: The probability that the algorithm will exit without finding any plane.
\item tolerance (distance): If a distance between an estimated plane and a point is larger than this tolerance, the point fits the estimated plane.
\item tolerance (angle): If an angle which consists of normal vectors of three randomly selected points and an estimated parameter $\vec n$ does not satisfy this tolerance, the selected points are regarded as unsuitable.
\item threshold: If the number of points that fits the estimated plane is bigger than this threshold, the points are regarded as a plane.
\end{itemize}
The threshold is selected as a variable in this experiment.
The experiment was conducted for ten values of the threshold and five environments.
The number of detected planes in each environment and each point density was recorded according to changing the threshold, where results of five trials were averaged.

\begin{figure}[tbp]    
	\centering
	\includegraphics[width=\columnwidth]{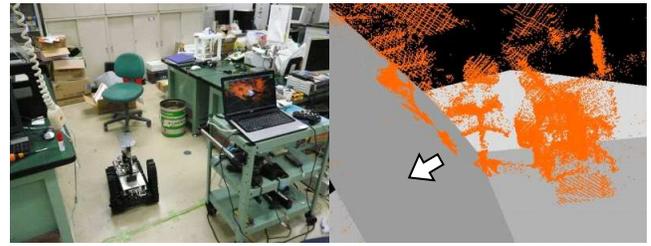}
	\caption{An example of a false detection for high point density in the case of threshold 5,000. Left: an actual environment. Right: a result of the plane detection. A nonexistent plane was detected as indicated by an arrow.}
	\label{FalseDetection}
\end{figure}

\begin{figure}[tbp]    
	\centering
	\subfigure[Environment 1]{\includegraphics[width=\columnwidth]{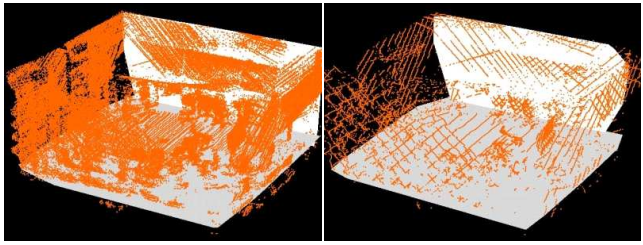}}
	\subfigure[Environment 2]{\includegraphics[width=\columnwidth]{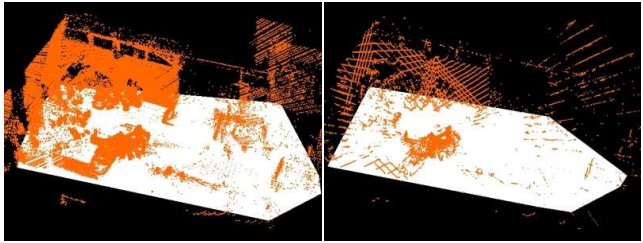}}
	\subfigure[Environment 3]{\includegraphics[width=\columnwidth]{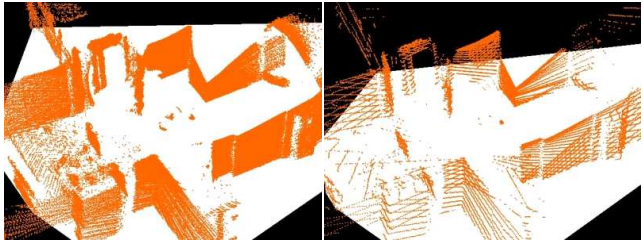}}
	\subfigure[Environment 4]{\includegraphics[width=\columnwidth]{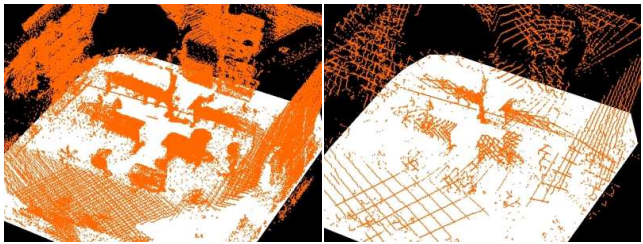}}
	\subfigure[Environment 5]{\includegraphics[width=\columnwidth]{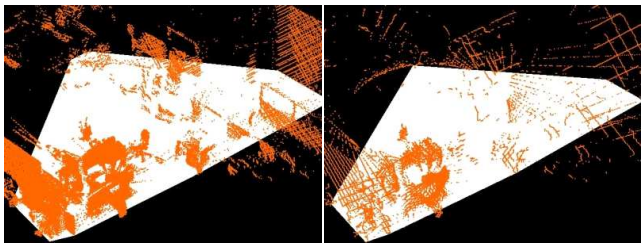}}	
	\caption{Examples of results in the case of threshold 15,000 for high point density data and threshold 3,000 for low point density data. Left: high point density. Right: low point density. Note that only the planes behind the point cloud are drawn and the points which compose planes are not displayed to be illustrated understandably.}
	\label{Environment_All}
\end{figure}

We checked with eyes whether there were false detections for every threshold.
False detections mean that planes were detected in void space where planes do not exist in the actual environment as shown in Fig. \ref{FalseDetection}.
As a result, we find that those false detections disappeared in the case of threshold more than 15,000 for high point density and more than 3,000 for low point density. 
Fig. \ref{Environment_All} shows results in the case of these thresholds.
Note that it is not adequate to choose the threshold in which the number of detected planes is maximized because it includes false detections.

\begin{figure}[tbp]    
	\centering
	\subfigure[Environment 1]{\includegraphics[width=\columnwidth]{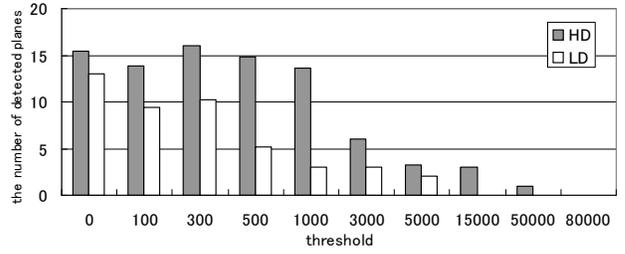}}
	\subfigure[Environment 2]{\includegraphics[width=\columnwidth]{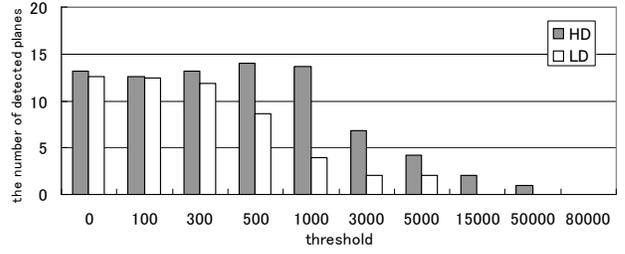}}
	\subfigure[Environment 3]{\includegraphics[width=\columnwidth]{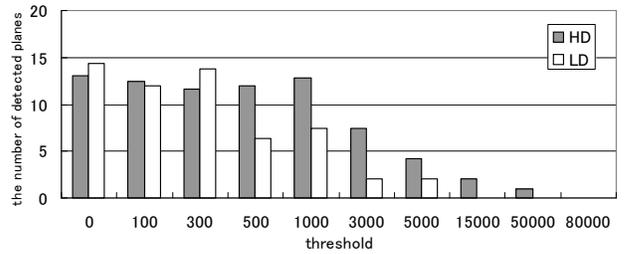}}
	\subfigure[Environment 4]{\includegraphics[width=\columnwidth]{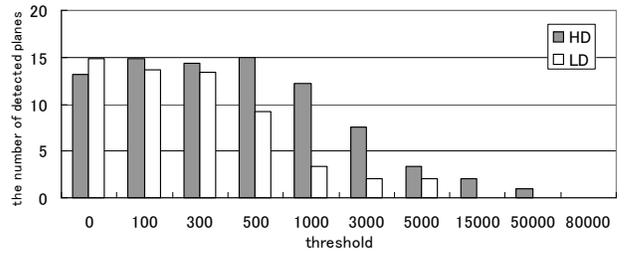}}
	\subfigure[Environment 5]{\includegraphics[width=\columnwidth]{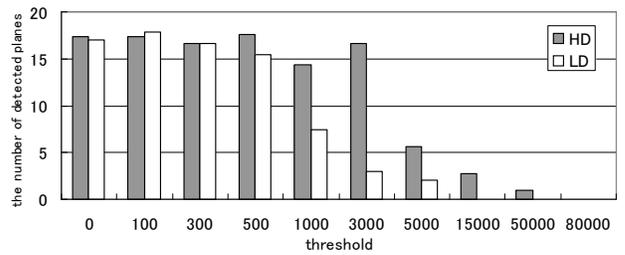}}	
	\caption{Results of the number of detected planes according to the threshold, where results of five trials were averaged. Each graph includes high and low point density data, which are indicated by HD and LD.}
	\label{Result_All_Graph}
\end{figure}

Fig. \ref{Result_All_Graph} shows graphs of results for five environments with high and low point density (indicated by HD and LD in graphs).
The number of detected planes was approximately constant 
in small thresholds. 
The number of detected planes was gradually decreased after beyond a certain threshold.

Any plane was not detected in the case of threshold 80,000 for high point density and threshold 15,000 for low point density. 

We conclude that the threshold of RANSAC algorithm should be changed according to point density in order to detect planes without false detections.
We suppose that the ratio of thresholds for high and low point density relates to the ratio of their point density ($6:1$) from the results of this experiment.

\section{Conclusion}
In this paper, we conducted an experiment to evaluate the plane detection method using RANSAC algorithm by changing its threshold and environments in both high and low point density data.
As a result, the threshold of RANSAC algorithm should be changed according to point density in order to detect planes without false detections.
In other words, low point density data is enough to detect planes by changing the threshold appropriately.
This result contributes to improve 3D scanning speed.

The future work is to conduct further experiments by changing the other parameters including probabilities $p_g$ and $p_{fail}$, and tolerances of distances and angles.

\bibliographystyle{junsrt}
\bibliography{reference.bib}

\begin{thebibliography}{1}

\bibitem{surmann2003autonomous}
Hartmut Surmann, Andreas N{\"u}chter, and Joachim Hertzberg.
\newblock An autonomous mobile robot with a {3D} laser range finder for {3D}
  exploration and digitalization of indoor environments.
\newblock {\em Robotics and Autonomous Systems}, Vol.~45, No.~3, pp. 181--198,
  2003.

\bibitem{fujiwara2013plane}
Tomofumi Fujiwara, Tetsushi Kamegawa, and Akio Gofuku.
\newblock Plane detection to improve 3d scanning speed using ransac algorithm.
\newblock In {\em Industrial Electronics and Applications (ICIEA), 2013 8th
  IEEE Conference on}, pp. 1863--1869, 2013.

\bibitem{fischler1981random}
Martin~A. Fischler and Robert~C. Bolles.
\newblock Random sample consensus: A paradigm for model fitting with
  applications to image analysis and automated cartography.
\newblock {\em Communications of the ACM}, Vol.~24, No.~6, pp. 381--395, 1981.

\bibitem{ransac_web}
Robert~B. Fisher.
\newblock The {RANSAC} (random sample consensus) algorithm.
\newblock
  http://homepages.inf.ed.ac.uk{\slash}rbf{\slash}CVonline{\slash}LOCAL{\_}COP%
IES{\slash}FISHER{\slash}RANSAC{\slash}{\#}fischler, 2002.

\bibitem{graham1972efficient}
R.~L. Graham.
\newblock An efficient algorith for determining the convex hull of a finite
  planar set.
\newblock {\em Information processing letters}, Vol.~1, No.~4, pp. 132--133,
  1972.

\bibitem{journals/jfr/KamegawaSHUM11}
Tetsushi Kamegawa, Noritaka Sato, Michinori Hatayama, Yojiro Uo, and Fumitoshi
  Matsuno.
\newblock Design and implementation of grouped rescue robot system using
  self-deploy networks.
\newblock {\em J. Field Robotics}, Vol.~28, No.~6, pp. 977--988, 2011.

\bibitem{ohno2009development}
Kazunori Ohno, Toyokazu Kawahara, and Satoshi Tadokoro.
\newblock Development of {3D} laser scanner for measuring uniform and dense
  {3D} shapes of static objects in dynamic environment.
\newblock In {\em Proceedings of the IEEE International Conference on Robotics
  and Biomimetics (ROBIO)}, pp. 2161--2167, 2009.

\bibitem{nagatani2008continuous}
Keiji Nagatani, Naoki Tokunaga, Yoshito Okada, and Kazuya Yoshida.
\newblock Continuous acquisition of three-dimensional environment information
  for tracked vehicles on uneven terrain.
\newblock In {\em Proceedings of the IEEE International Workshop on Safety,
  Security and Rescue Robotics (SSRR)}, pp. 25--30, 2008.

\end{thebibliography}

\end{document}